\documentclass[runningheads]{llncs}
\usepackage{graphicx}

\usepackage{tikz}
\usepackage{comment}
\usepackage{amsmath,amssymb} 
\usepackage{color}

\usepackage{hyperref}
\hypersetup{
    colorlinks=true,
    urlcolor=magenta,
}
\usepackage{subfigure}
\usepackage{multirow}
\usepackage{booktabs}
\usepackage{algorithm}
\usepackage{algorithmicx}
\usepackage[noend]{algpseudocode}
\algrenewcommand\algorithmicrequire{\textbf{Input:}}
\algrenewcommand\algorithmicensure{\textbf{Output:}}

\begin{document}
\pagestyle{headings}
\mainmatter
\def\ECCVSubNumber{2335}  

\title{Dynamic R-CNN: Towards High Quality Object Detection via Dynamic Training} 

\titlerunning{Dynamic R-CNN}
%
\author{Hongkai Zhang\inst{1,2} \and
Hong Chang\inst{1,2} \and
Bingpeng Ma\inst{2} \and\\
Naiyan Wang\inst{3} \and
Xilin Chen\inst{1,2}}
%
%
\institute{Key Laboratory of Intelligent Information Processing of Chinese Academy of Sciences (CAS), Institute of Computing Technology, Chinese Academy of Sciences \and
University of Chinese Academy of Sciences, \qquad $^3$ TuSimple\\
\email{hongkai.zhang@vipl.ict.ac.cn}, \email{changhong@ict.ac.cn}, \email{bpma@ucas.ac.cn}, \email{winsty@gmail.com}, \email{xlchen@ict.ac.cn}
}
\maketitle

\begin{abstract}
    Although two-stage object detectors have continuously advanced the state-of-the-art performance in recent years, the training process itself is far from crystal. In this work, we first point out the inconsistency problem between the fixed network settings and the dynamic training procedure, which greatly affects the performance. For example, the fixed label assignment strategy and regression loss function cannot fit the distribution change of proposals and thus are harmful to training high quality detectors.
    Consequently, we propose \emph{Dynamic R-CNN} to adjust the label assignment criteria (IoU threshold) and the shape of regression loss function (parameters of SmoothL1 Loss) automatically based on the statistics of proposals during training. This dynamic design makes better use of the training samples and pushes the detector to fit more high quality samples. Specifically, our method improves upon ResNet-50-FPN baseline with 1.9\% AP and 5.5\% AP$_{90}$ on the MS COCO dataset with no extra overhead. Codes and models are available at \url{https://github.com/hkzhang95/DynamicRCNN}.
    \keywords{dynamic training, high quality object detection}
\end{abstract}

\section{Introduction}

Benefiting from the advances in deep convolutional neural networks (CNNs) \cite{AlexNet,VGG,ResNet,AP3D}, object detection has made remarkable progress in recent years. Modern detection frameworks can be divided into two major categories of one-stage detectors~\cite{YOLO,SSD,FocalLoss} and two-stage detectors~\cite{RCNN,FastRCNN,FasterRCNN}. And various improvements have been made in recent studies~\cite{FCOS,TridentNet,RepPoints,CascadeRetinaNet,NoisyAnchors,CornerNet,RFBNet,AlignDet,LSTS}.
In the training procedure of both kinds of pipelines, a classifier and a regressor are adopted respectively to solve the \textit{recognition} and \textit{localization} tasks. Therefore, an effective training process plays a crucial role in achieving high quality object detection\footnote[1]{Specifically, high quality represents the results under high IoU.}.

\begin{figure}[!t]
    \centering
    \subfigure[]{
        \begin{minipage}{0.47\linewidth}
            \centering
            \includegraphics[width=\linewidth]{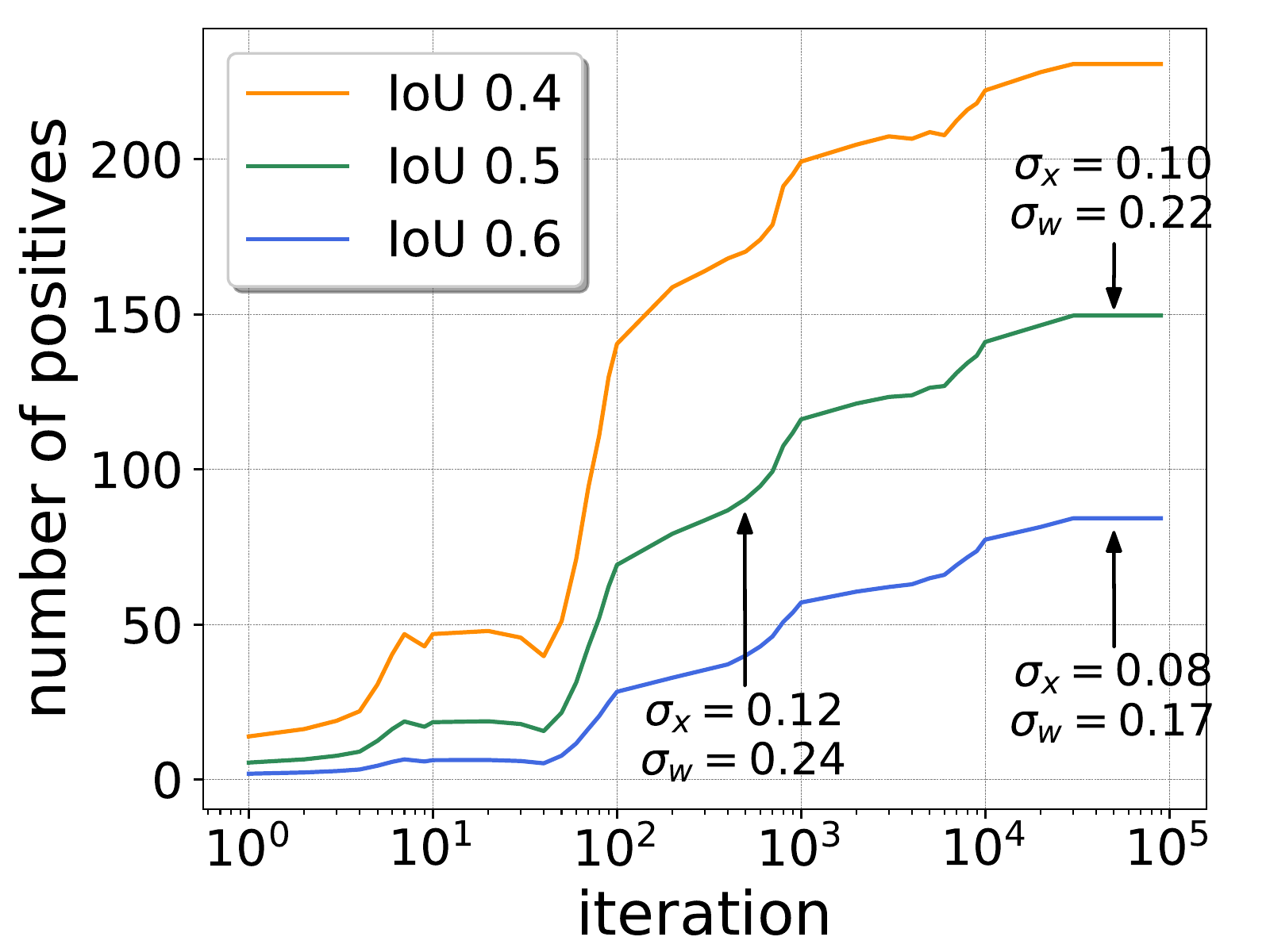}
        \end{minipage}
    }
    \subfigure[]{
        \begin{minipage}{0.47\linewidth}
            \centering
            \includegraphics[width=\linewidth]{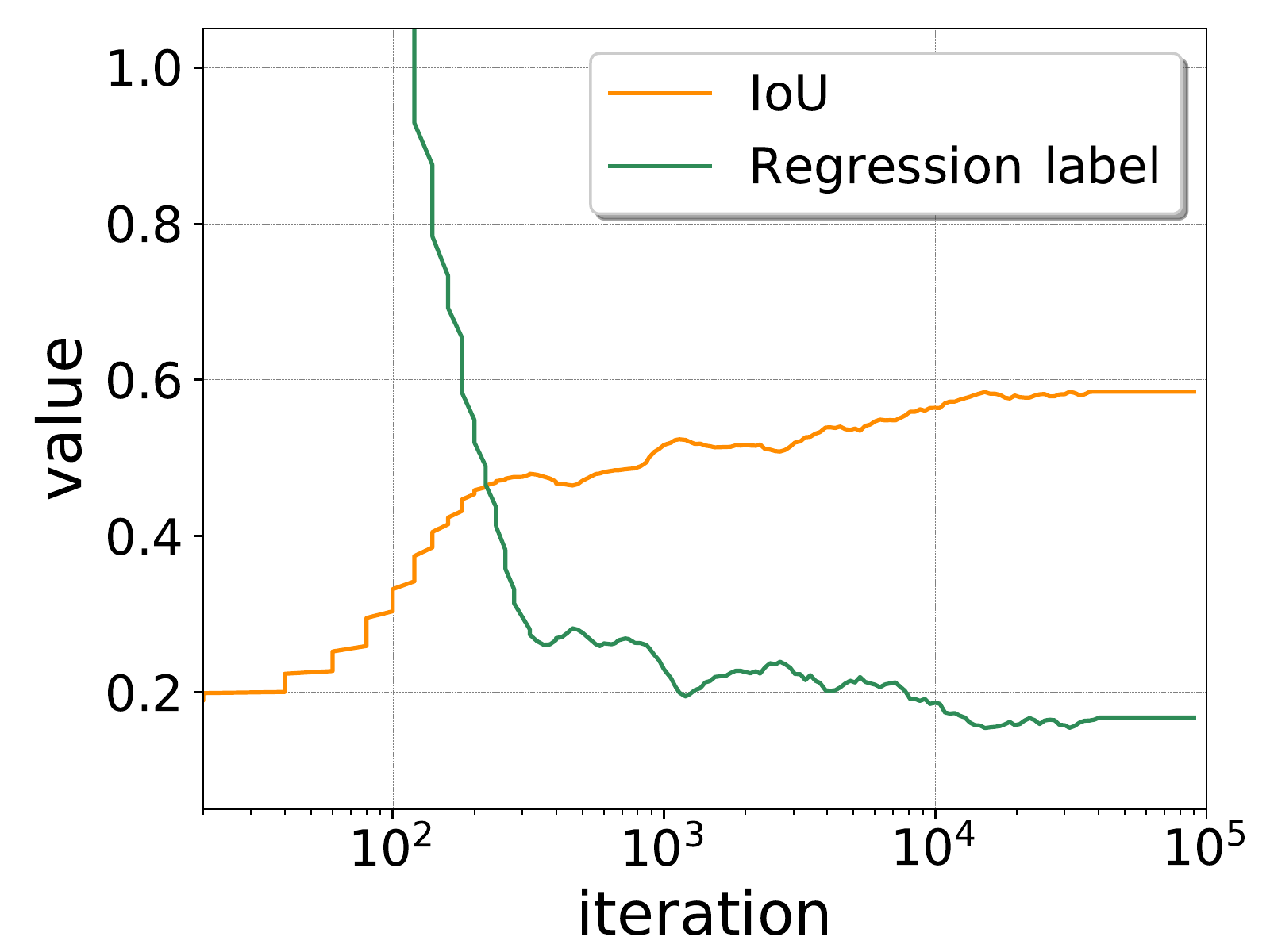}
        \end{minipage}
    }
    \caption{(a) The number of positive proposals under different IoU thresholds during the training process. The curve shows the numbers of positives vary significantly during training, with corresponding changes in regression labels distribution ($\sigma_x$ and $\sigma_w$ stands for the standard deviation for $x$ and $w$ respectively). (b) The IoU and regression label of the 75th and 10th most accurate proposals respectively in the training procedure. These curves further show the improved quality of proposals.
    }
    \label{fig:quantity}
\end{figure}

Different from the image classification task, the annotations for the classification task in object detection are the ground-truth boxes in the image. So it is not clear how to assign positive and negative labels for the proposals in classifier training since their separation may be ambiguous. The most widely used strategy is to set a threshold for the IoU of the proposal and corresponding ground-truth. As mentioned in Cascade R-CNN~\cite{CascadeRCNN}, training with a certain IoU threshold will lead to a classifier that degrades the performance at other IoUs. However, we cannot directly set a high IoU from the beginning of the training due to the scarcity of positive samples. The solution that Cascade R-CNN provides is to gradually refine the proposals by several stages, which are effective yet time-consuming. As for regressor, the problem is similar. During training, the quality of proposals is improved, however the parameter in SmoothL1 Loss is fixed. Thus it leads to insufficient training for the high quality proposals.

To solve this issue, we first examine an overlooked fact that the quality of proposals is indeed improved during training as shown in Figure~\ref{fig:quantity}. We can find that even under different IoU thresholds, the number of positives still increases significantly.
Inspired by the illuminating observations, we propose \textit{Dynamic R-CNN}, a simple yet effective method to better exploit the dynamic quality of proposals for object detection. It consists of two components: \textit{Dynamic Label Assignment} and \textit{Dynamic SmoothL1 Loss}, which are designed for classification and regression branches, respectively.
First, to train a better classifier that is discriminative for high IoU proposals, we gradually adjust the IoU threshold for positive/negative samples based on the proposals distribution in the training procedure. Specifically, we set the threshold as the IoU of the proposal at a certain percentage since it can reflect the quality of the overall distribution.
For regression, we choose to change the shape of the regression loss function to adaptively fit the distribution change of regression label and ensure the contribution of high quality samples to training. In particular, we adjust the $\beta$ in SmoothL1 Loss based on the regression label distribution, since $\beta$ actually controls the magnitude of the gradient of small errors (shown in Figure~\ref{fig:smoothL1}).

By this dynamic scheme, we can not only alleviate the data scarcity issue at the beginning of the training, but also harvest the benefit of high IoU training.
These two modules explore different parts of the detector, thus could work collaboratively towards high quality object detection.
Furthermore, despite the simplicity of our proposed method, Dynamic R-CNN could bring consistent performance gains on MS COCO~\cite{COCO} with almost no extra computational complexity in training. \emph{And during the inference phase, our method does not introduce any additional overhead.}  Moreover, extensive experiments verify the proposed method could generalize to other baselines with stronger performance.

\section{Related Work}

\textbf{Region-based object detectors.}
The general practice of region-based object detectors is converting the object detection task into a bounding box classification and a regression problem. In recent years, region-based approaches have been the leading paradigm with top performance. For example, R-CNN~\cite{RCNN}, Fast R-CNN~\cite{FastRCNN} and Faster R-CNN~\cite{FasterRCNN} first generate some candidate region proposals, then randomly sample a small batch with certain foreground-background ratio from all the proposals. These proposals will be fed into a second stage to classify the categories and refine the locations at the same time. Later, some works extended Faster R-CNN to address different problems. R-FCN~\cite{R-FCN} makes the whole network fully convolutional to improve the speed; and FPN~\cite{FPN} proposes a top-down pathway to combine multi-scale features. Besides, various improvements have been witnessed in recent studies~\cite{Speed-Accuracy-Tradeoff,GuidedAnchor,TridentNet,DetNet,DCNv2}.

\noindent
\textbf{Classification in object detection}.
Recent researches focus on improving object classifier from various perspectives~\cite{FocalLoss,IoUNet,LibraRCNN,LTR,NoisyAnchors,ATSS,DCR,KL-Loss}.
The classification scores in detection not only determine the semantic category for each proposal, but also imply the localization accuracy,
since Non-Maximum Suppression (NMS) suppresses less confident boxes using more reliable ones. It ranks the resultant boxes first using the classification scores.
However, as mentioned in IoU-Net~\cite{IoUNet}, the classification score has low correlation with localization accuracy, which leads to noisy ranking and limited performance. Therefore, IoU-Net~\cite{IoUNet} adopts an extra branch for predicting IoU scores and refining the classification confidence. Softer NMS~\cite{KL-Loss} devises an KL loss to model the variance of bounding box regression directly, and uses that for voting in NMS.
Another direction to improve is to raise the IoU threshold for training high quality classifiers, since training with different IoU thresholds will lead to classifiers with corresponding quality. However, as mentioned in Cascade R-CNN~\cite{CascadeRCNN}, directly raising the IoU threshold is impractical due to the vanishing positive samples. Therefore, to produce high quality training samples, some approaches~\cite{CascadeRCNN,CascadeRetinaNet} adopt sequential stages which are effective yet time-consuming. Essentially, it should be noted that these methods ignore the inherent dynamic property in training procedure which is useful for training high quality classifiers.

\noindent
\textbf{Bounding box regression.}
It has been proved that the performance of models is dependent on the relative weight between losses in multi-task learning~\cite{Multi-task}. Cascade R-CNN~\cite{CascadeRCNN} also adopt different regression normalization factors to adjust the aptitude of regression term in different stages. Besides, Libra R-CNN~\cite{LibraRCNN} proposes to promote the regression gradients from the accurate samples; and SABL~\cite{SABL} localizes each side of the bounding box with a lightweight two step bucketing scheme for precise localization. However, they mainly focus on a fixed scheme ignoring the dynamic distribution of learning targets during training.

\noindent
\textbf{Dynamic training.}
There are various researches following the idea of dynamic training. A widely used example is adjusting the learning rate based on the training iterations~\cite{SGDR}. Besides, Curriculum Learning~\cite{Curriculum} and Self-paced Learning~\cite{Self-paced} focus on improving the training order of the examples. Moreover, for object detection, hard mining methods~\cite{OHEM,FocalLoss,LibraRCNN} can also be regarded as a dynamic way. However, they don't handle the core issues in object detection such as constant label assignment strategy. Our method is complementary to theirs.

\section{Dynamic Quality in the Training Procedure}

Generally speaking, Object detection is complex since it needs to solve two main tasks: \textit{recognition} and \textit{localization}. \textit{Recognition} task needs to distinguish foreground objects from backgrounds and determine the semantic category for them. Besides, the \textit{localization} task needs to find accurate bounding boxes for different objects. To achieve high-quality object detection, we need to further explore the training process of both two tasks as follows.

\subsection{Proposal Classification}
\label{sec:analysis_cls}

\emph{How to assign labels} is an interesting question for the classifier in object detection. It is unique to other classification problems since the annotations are the ground-truth boxes in the image. Obviously, a proposal should be negative if it does not overlap with any ground-truth, and a proposal should be positive if its overlap with a ground-truth is 100\%. However, it is a dilemma to define whether a proposal with IoU 0.5 should be labeled as positive or negative.

In Faster R-CNN~\cite{FasterRCNN}, labels are assigned by comparing the box's highest IoU with ground-truths using a pre-defined IoU threshold. Formally, the paradigm can be formulated as follows (we take a binary classification loss for simplicity):

\begin{equation}
    \text{label} =
    \begin{cases}
        1, & \mbox{if } \max IoU(b, G) \geq T_+ \\
        0, & \mbox{if } \max IoU(b, G) < T_- \\
        -1, & \mbox{otherwise}.
    \end{cases}
    \label{eq:label_assign}
\end{equation}

Here $b$ stands for a bounding box, $G$ represents for the set of ground-truths, $T_+$ and $T_-$ are the positive and negative threshold for IoU. $1, 0, -1$ stand for positives, negatives and ignored samples, respectively. As for the second stage of Faster R-CNN, $T_+$ and $T_-$ are set to 0.5 by default~\cite{Detectron2018}. So the definition of positives and negatives is essentially hand-crafted.

Since the goal of classifier is to distinguish the positives and negatives, training with different IoU thresholds will lead to classifiers with corresponding quality~\cite{CascadeRCNN}.Therefore, to achieve high quality object detection, we need to train the classifier with a high IoU threshold. However, as mentioned in Cascade R-CNN, directly raising the IoU threshold is impractical due to the vanishing positive samples. Cascade R-CNN uses several sequential stages to lift the IoU of the proposals, which are effective yet time-consuming.

So is there a way to get the best of two worlds? As mentioned above, the quality of proposals actually improves along the training. This observation inspires us to take a progressive approach in training: At the beginning, the proposal network is not capable to produce enough high quality proposals, so we use a lower IoU threshold to better accommodate these imperfect proposals in second stage training. As training goes, the quality of proposals improves, we gradually have enough high quality proposals. As a result, we may increase the threshold to better utilize them to train a high quality detector that is more discriminative at higher IoU. We will formulate this process in the following section.

\subsection{Bounding Box Regression}
\label{sec:analysis_reg}

\begin{figure}[!t]
    \centering
    \includegraphics[width=0.9\linewidth]{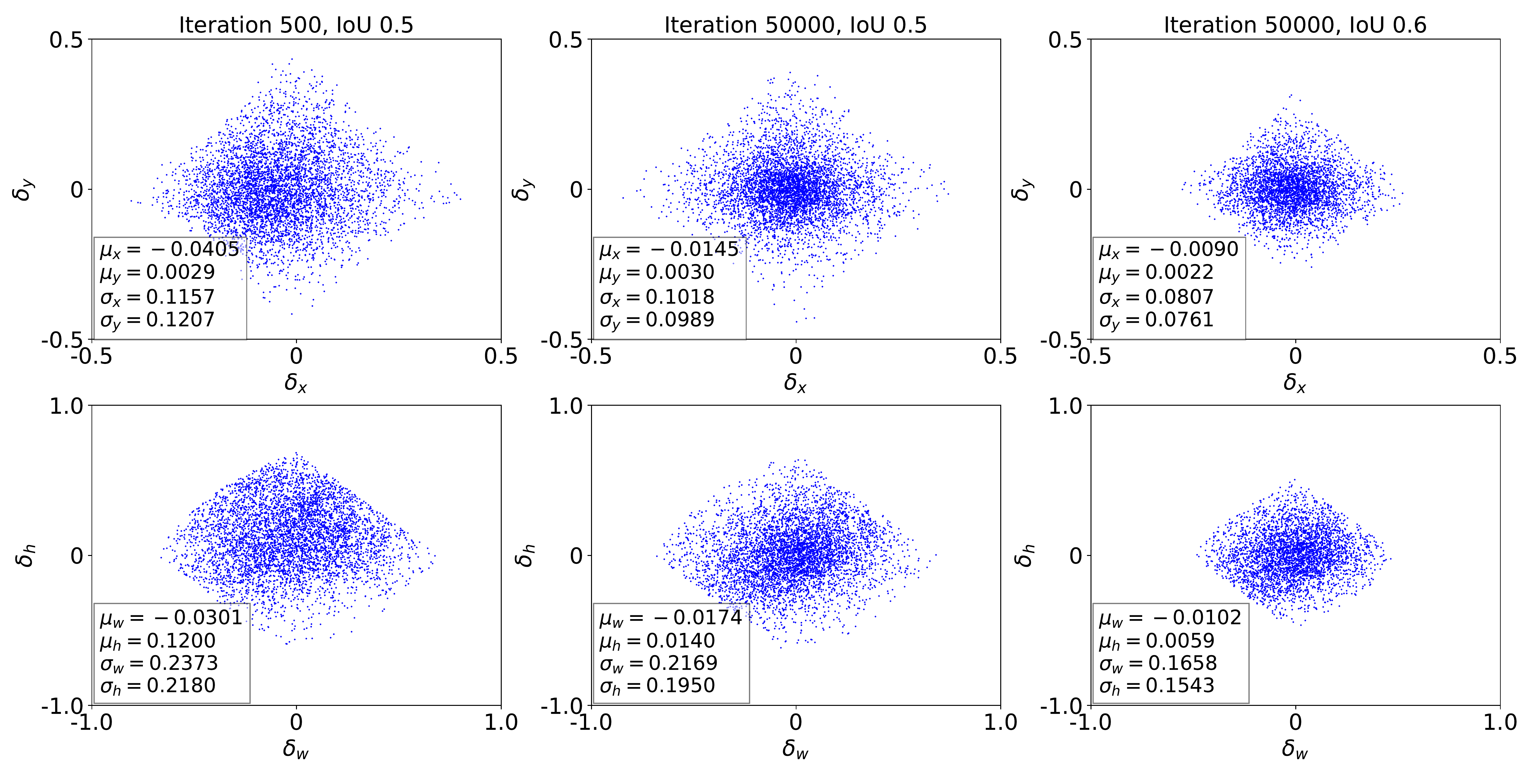}
    \caption{$\Delta$ distribution at different iterations and IoU thresholds (we randomly select some points for simplicity). Column 1\&2: under the same IoU threshold, the regression labels are more concentrated as the training goes. Column 2\&3: at the same iteration, raising the IoU threshold will significantly change the distribution.}
    \label{fig:reg_target_over_time}
\end{figure}

The task of bounding box regression is to regress the positive candidate bounding box $b$ to a target ground-truth $g$. This is learned under the supervision of the regression loss function $L_{reg}$. To encourage the regression label invariant to scale and location, $L_{reg}$ operates on the offset $\Delta=(\delta_x, \delta_y, \delta_w, \delta_h)$ defined by

\begin{equation}
    \begin{aligned}
        \delta_x=(g_x-b_x)/b_w, \quad \delta_y&=(g_y-b_y)/b_h\\
        \delta_w=\log(g_w/b_w), \quad \delta_h&=\log(g_h/b_h).
        \label{eq:offset}
    \end{aligned}
\end{equation}

Since the bounding box regression performs on the offsets, the absolute values of Equation~(\ref{eq:offset}) can be very small. To balance the different terms in multi-task learning, $\Delta$ is usually normalized by pre-defined \textit{mean} and \textit{stdev} (standard deviation) as widely used in many work~\cite{FasterRCNN,FPN,MaskRCNN}.

However, we discover that the distribution of regression labels are shifting during training. As shown in Figure~\ref{fig:reg_target_over_time}, we calculate the statistics of the regression labels under different iterations and IoU thresholds. First, from the first two columns, we find that under the same IoU threshold for positives, the \textit{mean} and \textit{stdev} are decreasing as the training goes due to the improved quality of proposals. With the same normalization factors, the contributions of those high quality samples will be reduced based on the definition of SmoothL1 Loss function, which is harmful to the training of high quality regressors.
Moreover, with a higher IoU threshold, the quality of positive samples is further enhanced, thus their contributions are reduced even more, which will greatly limit the overall performance.
Therefore, to achieve high quality object detection, we need to fit the distribution change and adjust the shape of regression loss function to compensate for the increasing of high quality proposals.

\section{Dynamic R-CNN}
\label{sec:dynamic_rcnn}

\begin{figure}[!t]
    \centering
    \includegraphics[width=0.9\linewidth]{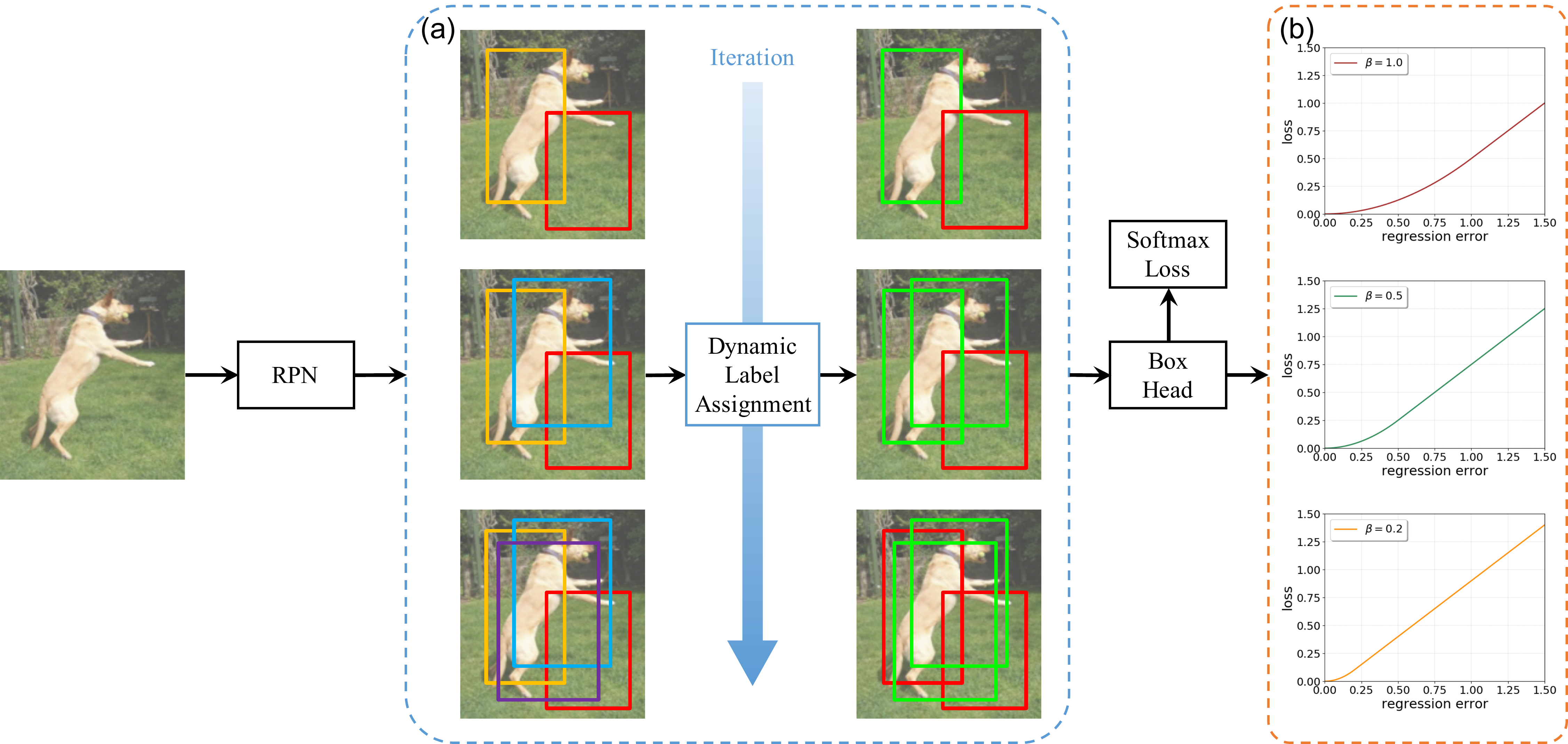}
    \caption{The overall pipeline of the proposed Dynamic R-CNN. Considering the dynamic property of the training process, Dynamic R-CNN consists of two main components (a) Dynamic Label Assignment (DLA) process and (b) Dynamic SmoothL1 Loss (DSL) from different perspectives. From the left part of (a) we can find that there are more high quality proposals as the training goes. With the improved quality of proposals, DLA will automatically raise the IoU threshold based on the proposal distribution. Then positive (green) and negative (red) labels are assigned for the proposals by DLA which are shown in the right part of (a). Meanwhile, to fit the distribution change and compensate for the increasing of high quality proposals, the shape of regression loss function is also adjusted correspondingly in (b). Best viewed in color.}
    \label{fig:pipeline}
\end{figure}

To better exploit the dynamic property of the training procedure, we propose Dynamic R-CNN which is shown in Figure~\ref{fig:pipeline}. Our key insight is \textbf{adjusting the second stage classifier and regressor to fit the distribution change of proposals}. The two components designed for the classification and localization branch will be elaborated in the following sections.

\subsection{Dynamic Label Assignment}

The Dynamic Label Assignment (DLA) process is illustrated in Figure~\ref{fig:pipeline} (a). Based on the common practice of label assignment in Equation~(\ref{eq:label_assign}) in object detection, the DLA module can be formulated as follows:

\begin{equation}
    \text{label} =
    \begin{cases}
        1, & \mbox{if } \max IoU(b, G) \geq T_{now}\\
        0, & \mbox{if } \max IoU(b, G) < T_{now},
    \end{cases}
    \label{eq:dynamic_label_assign}
\end{equation}
where $T_{now}$ stands for the current IoU threshold. Considering the dynamic property in training, the distribution of proposals is changing over time. Our DLA updates the $T_{now}$ automatically based on the statistics of proposals to fit this distribution change. Specifically, we first calculate the IoUs $I$ between proposals and their target ground-truths, and then select the $K_I$-th largest value from $I$ as the threshold $T_{now}$. As the training goes, $T_{now}$ will increase gradually which reflects the improved quality of proposals. In practice, we first calculate the $K_I$-th largest value in each batch, and then update $T_{now}$ every $C$ iterations using the mean of them to enhance the robustness of the training. It should be noted that the calculation of IoUs is already done by the original method, so there is almost no additional complexity in our method.
The resultant IoU thresholds used in training are illustrated in Figure~\ref{fig:pipeline} (a).

\subsection{Dynamic SmoothL1 Loss}

The localization task for object detection is supervised by the commonly used SmoothL1 Loss, which can be formulated as follows:

\begin{equation}
    SmoothL1(x, \beta) =
    \begin{cases}
        0.5|x|^2/\beta, & \mbox{if }|x|<\beta, \\
        |x|-0.5\beta, & \mbox{otherwise}.
    \end{cases}
    \label{eq:smoothL1_loss}
\end{equation}

\begin{figure}[!t]
    \centering
    \subfigure[]{
        \begin{minipage}{0.47\linewidth}
            \centering
            \includegraphics[width=\linewidth]{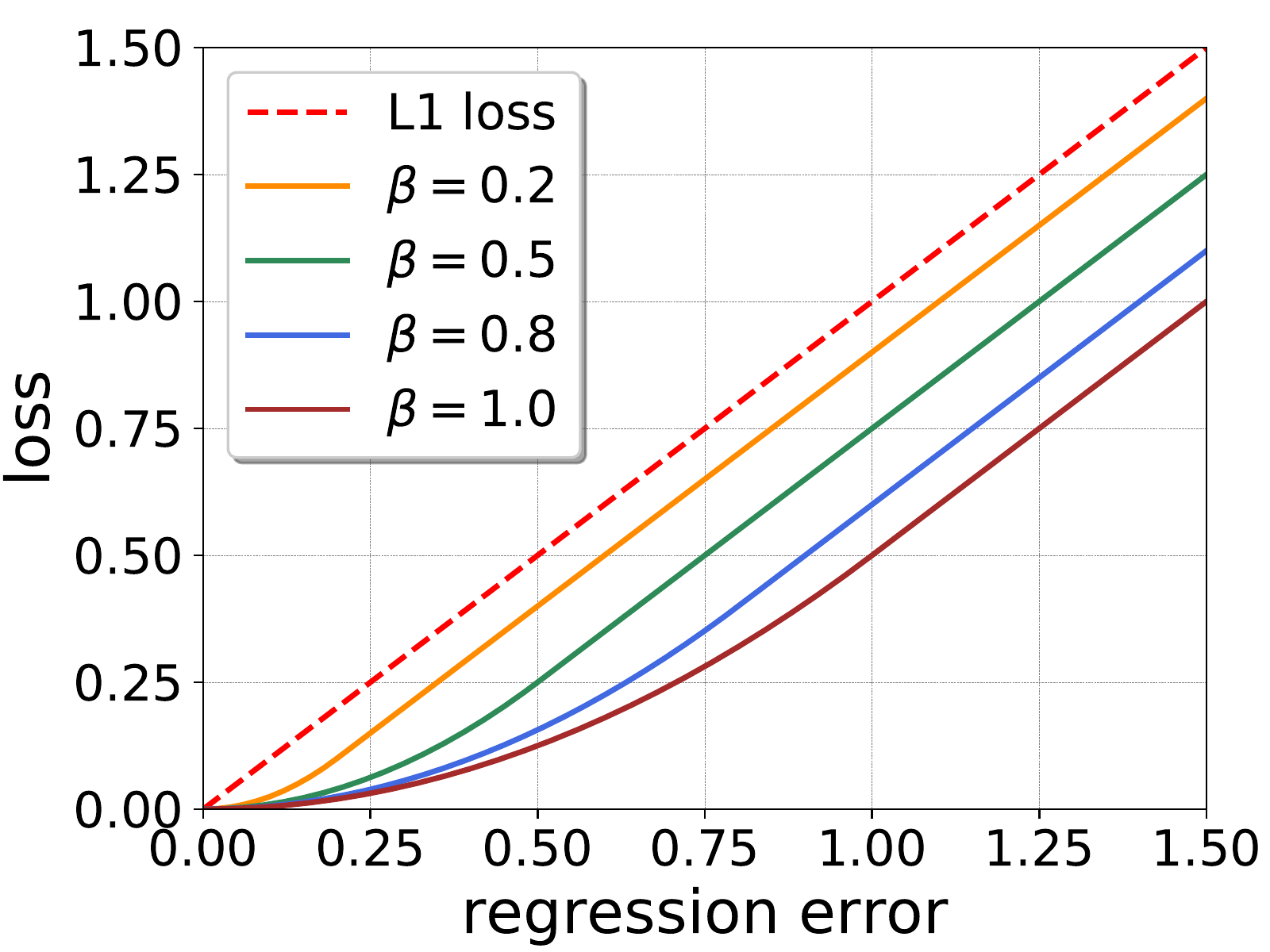}
        \end{minipage}
    }
    \subfigure[]{
        \begin{minipage}{0.47\linewidth}
            \centering
            \includegraphics[width=\linewidth]{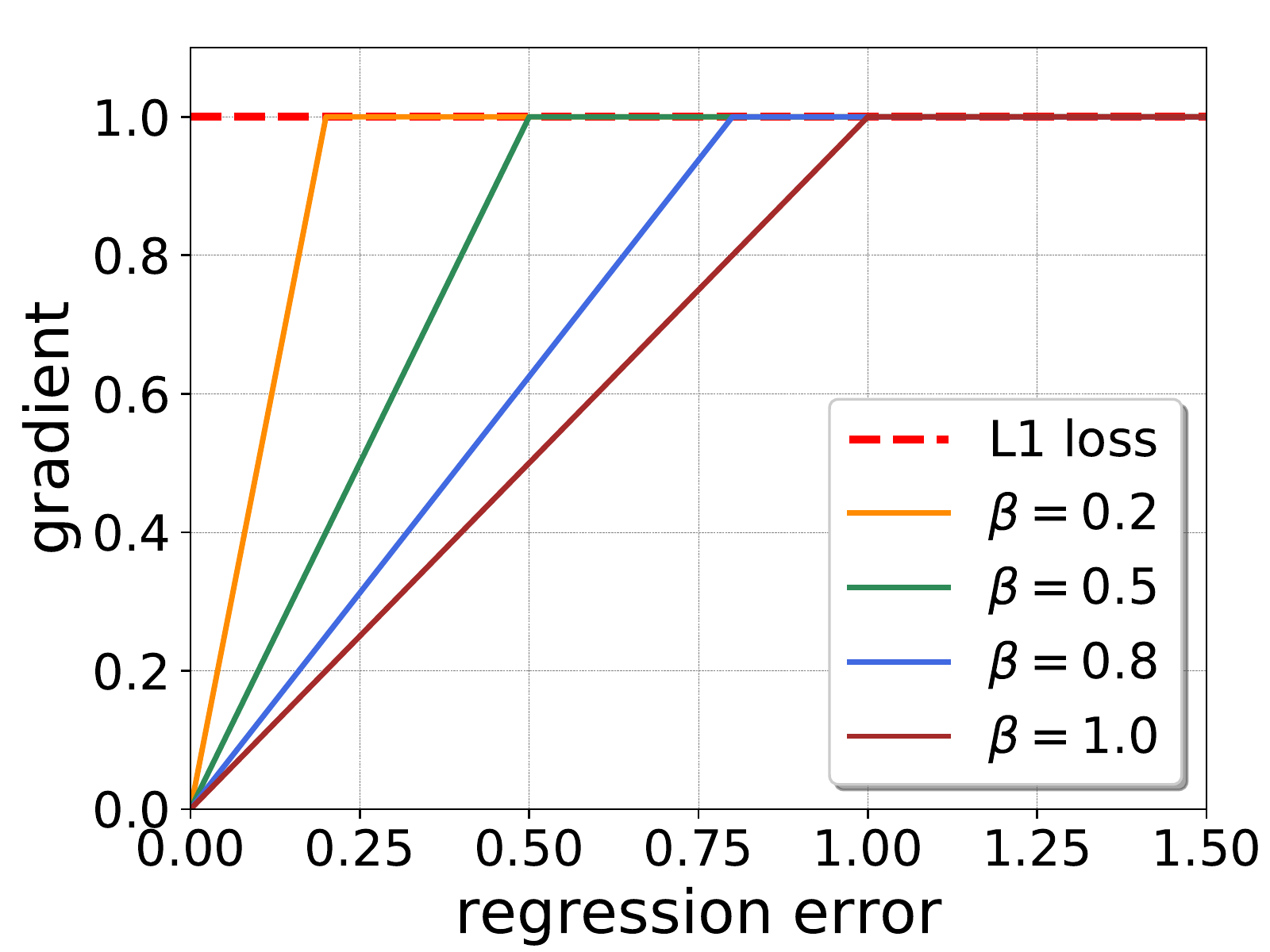}
        \end{minipage}
    }
    \caption{We show curves for (a) loss and (b) gradient of SmoothL1 Loss with different $\beta$ here. $\beta$ is set default as 1.0 in the R-CNN part.}
    \label{fig:smoothL1}
\end{figure}

Here the $x$ stands for the regression label. $\beta$ is a hyper-parameter controlling in which range we should use a softer loss function like $l_1$ loss instead of the original $l_2$ loss. Considering the robustness of training, $\beta$ is set default as 1.0 to prevent the exploding loss due to the poor trained network in the early stages. We also illustrate the impact of $\beta$ in Figure~\ref{fig:smoothL1}, in which changing $\beta$ leads to different curves of loss and gradient. It is easy to find that a smaller $\beta$ actually accelerate the saturation of the magnitude of gradient, thus it makes more accurate sample contributes more to the network training.

As analyzed in Section~\ref{sec:analysis_reg}, we need to fit the distribution change and adjust the regression loss function to compensate for the high quality samples. So we propose Dynamic SmoothL1 Loss (DSL) to change the shape of loss function to gradually focus on high quality samples as follows:

\begin{equation}
    DSL(x, \beta_{now}) =
    \begin{cases}
        0.5|x|^2/\beta_{now}, & \mbox{if }|x|<\beta_{now}, \\
        |x|-0.5\beta_{now}, & \mbox{otherwise}.
    \end{cases}
    \label{eq:DSL}
\end{equation}

Similar to DLA, DSL will change the value of $\beta_{now}$ according to the statistics of regression labels which can reflect the localization accuracy. To be more specific, we first obtain the regression labels $E$ between proposals and their target ground-truths, then select the $K_\beta$-th smallest value from $E$ to update the $\beta_{now}$ in Equation~(\ref{eq:smoothL1_loss}). Similarly, we also update the $\beta_{now}$ every $C$ iterations using the median of the $K_{\beta}$-th smallest label in each batch. We choose median instead of mean as in the classification because we find more outliers in regression labels.
Through this dynamic way, appropriate $\beta_{now}$ will be adopted automatically as shown in Figure~\ref{fig:pipeline} (b), which will better exploit the training samples and lead to a high quality regressor.

To summarize the whole method, we describe the proposed Dynamic R-CNN in Algorithm~\ref{algm:DynamicRCNN}. Besides the proposals $P$ and ground-truths $G$, Dynamic R-CNN has three hyperparamters: IoU threshold top-k $K_I$, $\beta$ top-k $K_\beta$ and update iteration count $C$. Note that compared with baseline, we only introduce one additional hyperparameter. And we will show soon the results are actually quite robust to the choice of these hyperparameters.

\begin{algorithm}[!t]
    \caption{Dynamic R-CNN}
    \begin{algorithmic}[1]
    \Require
        \Statex Proposal set $P$, ground-truth set $G$.
        \Statex IoU threshold top-k $K_{I}$, $\beta$ top-k $K_\beta$, update iteration count $C$.
    \Ensure
        \Statex Trained object detector $D$.
    \State Initialize IoU threshold and SmoothL1 $\beta$ as $T_{now}, \beta_{now}$
    \State Build two empty sets $\mathcal{S}_{I}, \mathcal{S}_{E}$ for recording the IoUs and regression labels
    \For {$i=0$ to max\_iter}
        \State Obtain matched IoUs $I$ and regression labels $E$ between $P$ and $G$
        \State Select thresholds $I_k, E_k$ based on the $K_{I}, K_\beta$
        \State Record corresponding values, add $I_k$ to $\mathcal{S}_{I}$ and $E_k$ to $\mathcal{S}_{E}$
        \If {$i\ \%\ C == 0$}
            \State Update IoU threshold: $T_{now}=\text{Mean}(\mathcal{S}_{I})$
            \State Update SmoothL1 $\beta$: $\beta_{now}=\text{Median}(\mathcal{S}_{E})$
            \State $\mathcal{S}_{I}=\text{\O}, \mathcal{S}_{E}=\text{\O}$
        \EndIf
        \State Train the network with $T_{now}, \beta_{now}$
    \EndFor
    \State \Return{Improved object detector $D$}
    \end{algorithmic}
    \label{algm:DynamicRCNN}
\end{algorithm}

\section{Experiments}

\subsection{Dataset and Evaluation Metrics}

Experimental results are mainly evaluated on the bounding box detection track of the challenging MS COCO~\cite{COCO} 2017 dataset. Following the common practice~\cite{FocalLoss,MaskRCNN}, we use the COCO \texttt{train} split ($\sim$118k images) for training and report the ablation studies on the \texttt{val} split (5k images). We also submit our main results to the evaluation server for the final performance on the \texttt{test-dev} split, \textit{which has no disclosed labels}. The COCO-style Average Precision (AP) is chosen as the main evaluation metric which averages AP across IoU thresholds from 0.5 to 0.95 with an interval of 0.05. We also include other metrics to better understand the behavior of the proposed method.

\begin{table}[!t]
    \caption{Comparisons with different baselines (our re-implementations) on COCO \texttt{test-dev} set. ``MST'' and ``*'' stand for multi-scale training and testing respectively. ``$2\times$'' and ``$3\times$'' are training schedules which extend the iterations by $2/3$ times.}
    \label{tab:main_comparison}
    \begin{center}
\setlength{\tabcolsep}{2pt}
\begin{tabular}{lccccccc}
\toprule
Method & Backbone & $\mathrm{AP}$ & $\mathrm{AP}_{50}$ & $\mathrm{AP}_{75}$ & $\mathrm{AP_S}$ & $\mathrm{AP_M}$ & $\mathrm{AP_L}$\\
\midrule
Faster R-CNN & ResNet-50 & 37.3 & 58.5 & 40.6 & 20.3 & 39.2 & 49.1\\
Faster R-CNN+$2\times$ & ResNet-50 & 38.1 & 58.9 & 41.5 & 20.5 & 40.0 & 50.0\\
Faster R-CNN & ResNet-101 & 39.3 & 60.5 & 42.7 & 21.3 & 41.8 & 51.7\\
Faster R-CNN+$2\times$ & ResNet-101 & 39.9 & 60.6 & 43.5 & 21.4 & 42.4 & 52.1\\
Faster R-CNN+$3\times$+MST & ResNet-101 & 42.8 & 63.8 & 46.8 & 24.8 & 45.6 & 55.6\\
Faster R-CNN+$3\times$+MST & ResNet-101-DCN & 44.8 & 65.5 & 48.8 & 26.2 & 47.6 & 58.1\\
Faster R-CNN+$3\times$+MST* & ResNet-101-DCN & 46.9 & 68.1 & 51.4 & 30.6 & 49.6 & 58.1\\
\midrule
Dynamic R-CNN & ResNet-50 & 39.1 & 58.0 & 42.8 & 21.3 & 40.9 & 50.3\\
Dynamic R-CNN+$2\times$ & ResNet-50 & 39.9 & 58.6 & 43.7 & 21.6 & 41.5 & 51.9\\
Dynamic R-CNN & ResNet-101 & 41.2 & 60.1 & 45.1 & 22.5 & 43.6 & 53.2\\
Dynamic R-CNN+$2\times$ & ResNet-101 & 42.0 & 60.7 & 45.9 & 22.7 & 44.3 & 54.3\\
Dynamic R-CNN+$3\times$+MST & ResNet-101 & 44.7 & 63.6 & 49.1 & 26.0 & 47.4 & 57.2\\
Dynamic R-CNN+$3\times$+MST & ResNet-101-DCN & 46.9 & 65.9 & 51.3 & 28.1 & 49.6 & 60.0\\
Dynamic R-CNN+$3\times$+MST* & ResNet-101-DCN & 49.2 & 68.6 & 54.0 & 32.5 & 51.7 & 60.3\\
\bottomrule
\end{tabular}
\end{center}

\end{table}

\subsection{Implementation Details}

For fair comparisons, all experiments are implemented on PyTorch~\cite{PyTorch} and follow the settings in maskrcnn-benchmark\footnote[2]{\url{https://github.com/facebookresearch/maskrcnn-benchmark}} and SimpleDet~\cite{SimpleDet}. We adopt FPN-based Faster R-CNN~\cite{FasterRCNN,FPN} with ResNet-50~\cite{ResNet} model pre-trained on ImageNet~\cite{ImageNet} as our baseline. All models are trained on the COCO 2017 \texttt{train} set and tested on \texttt{val} set with image short size at 800 pixels unless noted. Due to the scarcity of positives in the training procedure, we set the NMS threshold of RPN to 0.85 instead of 0.7 for all the experiments.

\subsection{Main Results}

We compare Dynamic R-CNN with corresponding baselines on COCO \texttt{test-dev} set in Table~\ref{tab:main_comparison}. For fair comparisons, We report our re-implemented results.

First, we prove that our method can work on different backbones. Dynamic R-CNN achieves 39.1\% AP with ResNet-50~\cite{ResNet}, which is 1.8 points higher than the FPN-based Faster R-CNN baseline. With a stronger backbone like ResNet-101, Dynamic R-CNN can also achieve consistent gains (+1.9 points).

Then, our dynamic design is also compatible with other training and testing skills.
The results are consistently improved by progressively adding in $2\times$ longer training schedule, multi-scale training (extra $1.5\times$ longer training schedule), multi-scale testing and deformable convolution~\cite{DCNv2}. With the best combination, out Dynamic R-CNN achieves 49.2\% AP, which is still 2.3 points higher than the Faster R-CNN baseline.

These results show the effectiveness and robustness of our method since it can work together with different backbones and multiple training and testing skills. It should also be noted that the performance gains are almost free.

\begin{table}[!t]
    \caption{Results of each component in Dynamic R-CNN on COCO \texttt{val} set.}
    \label{tab:components}
    \begin{center}
\setlength{\tabcolsep}{4pt}
\begin{tabular}{cccccccccc}
\toprule
Backbone & DLA & DSL & $\mathrm{AP}$ & $\Delta\mathrm{AP}$ & $\mathrm{AP}_{50}$ & $\mathrm{AP}_{60}$ & $\mathrm{AP}_{70}$ & $\mathrm{AP}_{80}$ & $\mathrm{AP}_{90}$\\
\midrule
ResNet-50-FPN & & & 37.0 & - & 58.0 & 53.5 & 46.0 & 32.6 & 9.7\\
ResNet-50-FPN & & \checkmark & 38.0 & +1.0 & 57.6 & 53.5 & 46.7 & 34.4 & 13.2\\
ResNet-50-FPN & \checkmark &  & 38.2 & +1.2 & 57.5 & 53.6 & 47.1 & 35.2 & 12.6\\
ResNet-50-FPN & \checkmark & \checkmark & 38.9 & +1.9 & 57.3 & 53.6 & 47.4 & 36.3 & 15.2\\
\bottomrule
\end{tabular}
\end{center}

\end{table}

\begin{figure}[!t]
    \centering
    \subfigure[]{
        \begin{minipage}{0.47\linewidth}
            \centering
            \includegraphics[width=\linewidth]{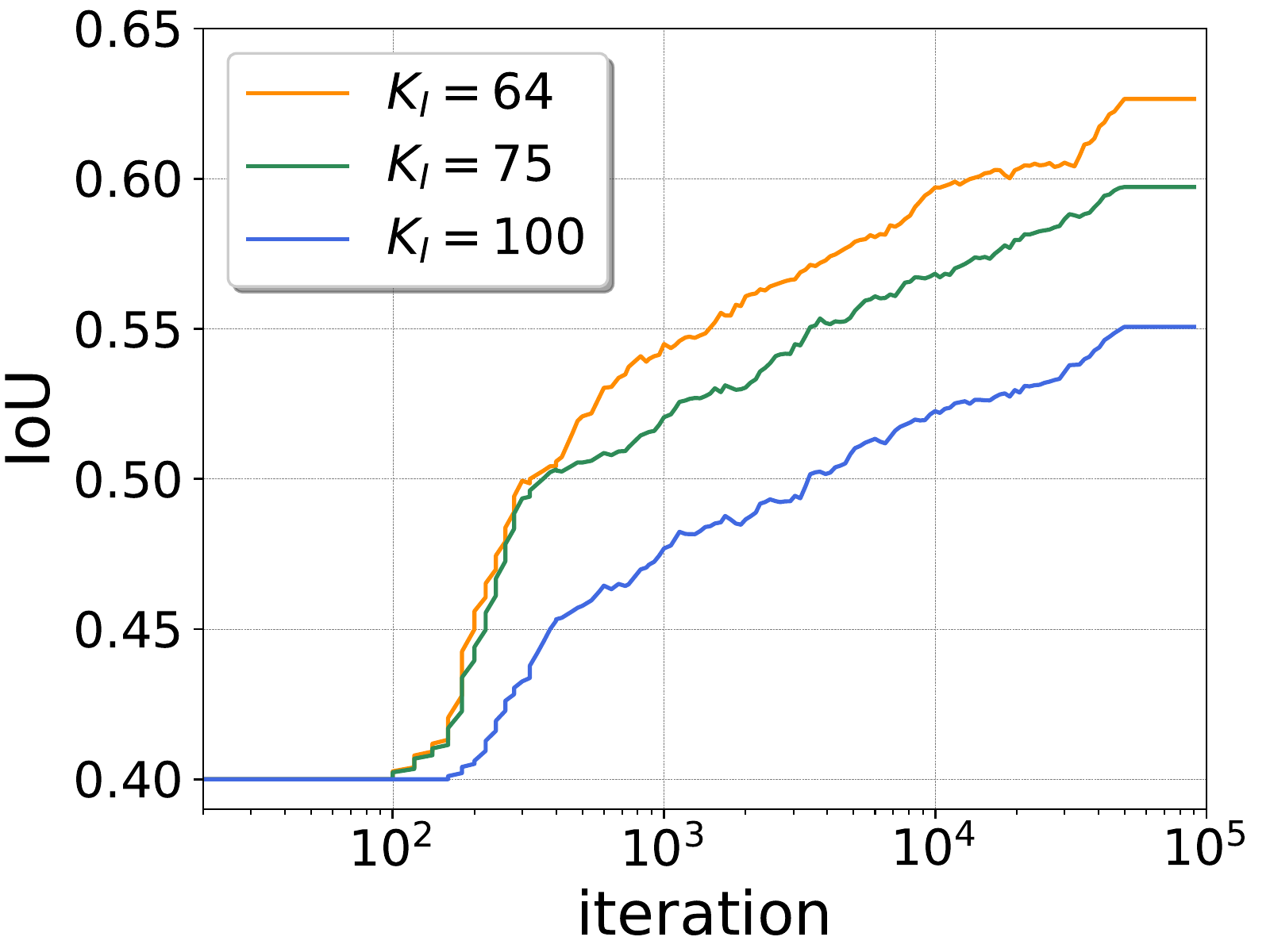}
        \end{minipage}
    }
    \subfigure[]{
        \begin{minipage}{0.47\linewidth}
            \centering
            \includegraphics[width=\linewidth]{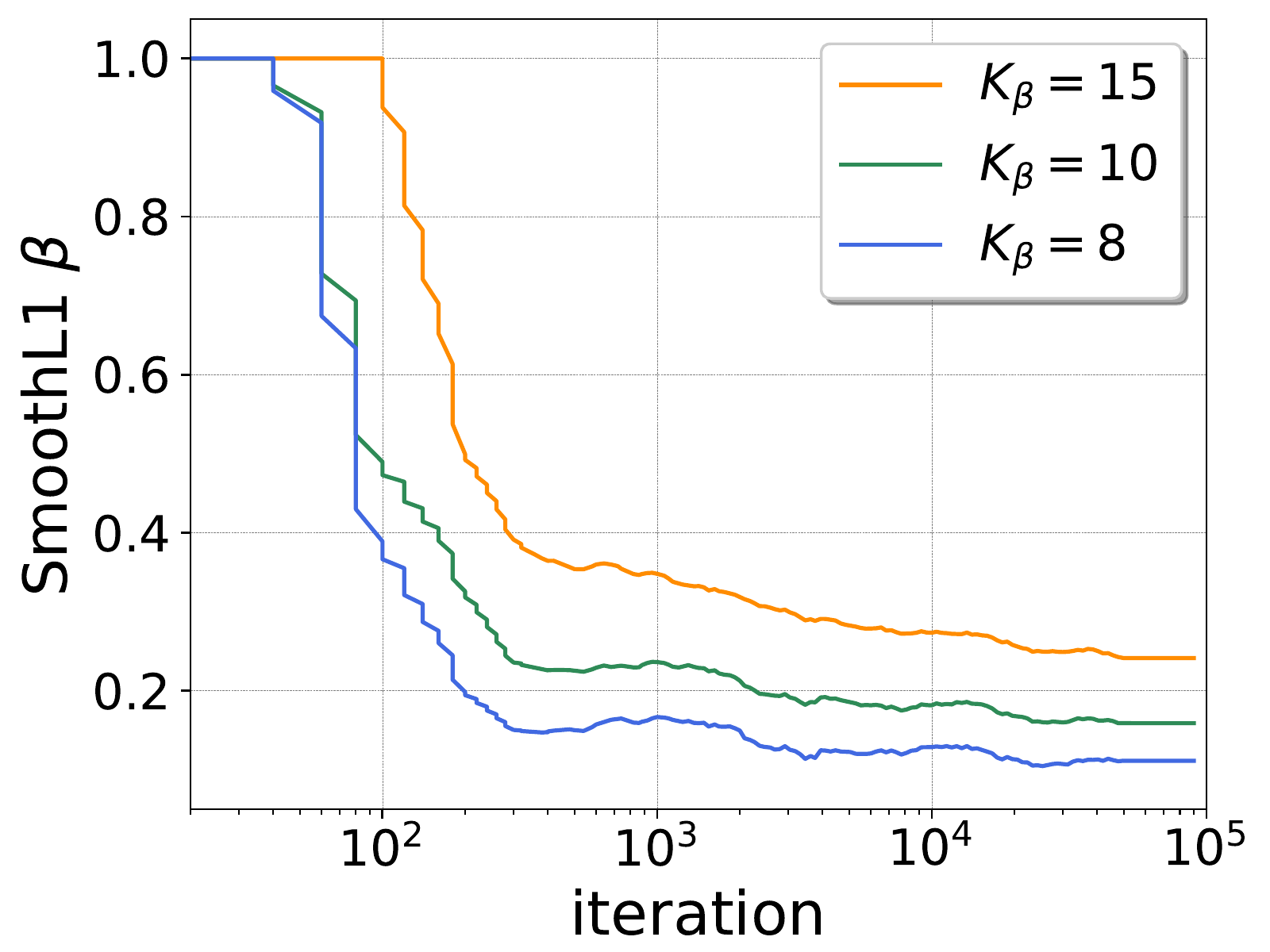}
        \end{minipage}
    }
    \caption{Trends of (a) IoU threshold and (b) SmoothL1 $\beta$ under different settings based on our method. Obviously the distribution has changed a lot during training.}
    \label{fig:dynamic_trend}
\end{figure}

\subsection{Ablation Experiments}
To show the effectiveness of each proposed component, we report the overall ablation studies in Table~\ref{tab:components}.

\noindent
\textbf{1) Dynamic Label Assignment (DLA).}
DLA brings 1.2 points higher box AP than the ResNet-50-FPN baseline. To be more specific, results in higher IoU metrics are consistently improved, especially for the 2.9 points gains in $\mathrm{AP_{90}}$. It proves the effectiveness of our method for pushing the classifier to be more discriminative at higher IoU thresholds.

\noindent
\textbf{2) Dynamic SmoothL1 Loss (DSL).}
DSL improves the box AP from 37.0 to 38.0. Results in higher IoU metrics like $\mathrm{AP}_{80}$ and $\mathrm{AP}_{90}$ are hugely improved, which validates the effectiveness of changing the loss function to compensate for the high quality samples during training. Moreover, as analyzed in Section~\ref{sec:analysis_reg}, with DLA the quality of positives is further improved thus their contributions are reduced even more. So applying DSL on DLA will also bring reasonable gains especially on high quality metrics. \textbf{To sum up, Dynamic R-CNN improves the baseline by 1.9 points $\mathrm{AP}$ and 5.5 points $\mathrm{AP_{90}}$}.

\noindent
\textbf{3) Illustration of dynamic training.}
To further illustrate the dynamics in the training procedure, we show the trends of IoU threshold and SmoothL1 $\beta$ under different settings based on our method in Figure~\ref{fig:dynamic_trend}. Here we clip the values of IoU threshold and $\beta$ to 0.4 and 1.0 respectively at the beginning of training.
Regardless of the specific values of $K_I$ and $K_\beta$, the overall trend of IoU threshold is increasing while that for SmoothL1 $\beta$ is decreasing during training. These results again verify the proposed method work as expected.

\noindent
\begin{minipage}[!t]{\textwidth}
    \begin{minipage}{0.49\textwidth}
    \centering
    \makeatletter\def\@captype{table}\makeatother\caption{Ablation study on $K_I$.}
        \setlength{\tabcolsep}{1pt}
        \begin{tabular}{ccccccc}
        \toprule
        $K_I$ & $\mathrm{AP}$ & $\mathrm{AP}_{50}$ & $\mathrm{AP}_{60}$ & $\mathrm{AP}_{70}$ & $\mathrm{AP}_{80}$ & $\mathrm{AP}_{90}$\\
        \midrule
        - & 37.0 & 58.0 & 53.5 & 46.0 & 32.6 & 9.7\\
        \midrule
        64 & 38.1 & 57.2 & 53.3 & 46.8 & 35.1 & \textbf{12.8}\\
        75 & \textbf{38.2} & 57.5 & 53.6 & \textbf{47.1} & \textbf{35.2} & 12.6\\
        100 & 37.9 & \textbf{57.9} & \textbf{53.8} & 46.9 & 34.2 & 11.6\\
        \bottomrule
        \end{tabular}
        \label{tab:dynamic_label_assignment}
    \end{minipage}
    \begin{minipage}{0.5\textwidth}
    \centering
        \makeatletter\def\@captype{table}\makeatother\caption{Ablation study on $C$.}
        \setlength{\tabcolsep}{1pt}
        \begin{tabular}{ccccccc}
        \toprule
        $C$ & $\mathrm{AP}$ & $\mathrm{AP}_{50}$ & $\mathrm{AP}_{60}$ & $\mathrm{AP}_{70}$ & $\mathrm{AP}_{80}$ & $\mathrm{AP}_{90}$\\
        \midrule
        - & 37.0 & 58.0 & 53.5 & 46.0 & 32.6 & 9.7\\
        \midrule
        20 & 38.0 & 57.4 & 53.5 & 47.0 & 35.0 & 12.5\\
        100 & 38.2 & 57.5 & 53.6 & 47.1 & 35.2 & 12.6\\
        500 & 38.1 & 57.6 & 53.5 & 47.2 & 34.8 & 12.6\\
        \bottomrule
        \end{tabular}
        \label{tab:iteration_count}
    \end{minipage}
\end{minipage}

\noindent
\begin{minipage}[!t]{\textwidth}
    \begin{minipage}{0.56\textwidth}
    \centering
        \makeatletter\def\@captype{table}\makeatother\caption{Ablation study on $K_\beta$.}
        \setlength{\tabcolsep}{1pt}
        \begin{tabular}{ccccccc}
        \toprule
        Setting & $\mathrm{AP}$ & $\mathrm{AP}_{50}$ & $\mathrm{AP}_{60}$ & $\mathrm{AP}_{70}$ & $\mathrm{AP}_{80}$ & $\mathrm{AP}_{90}$\\
        \midrule
        $\beta=1.0$ & 37.0 & 58.0 & 53.5 & 46.0 & 32.6 & 9.7\\
        $\beta=2.0$ & 35.9 & 57.7 & 53.2 & 45.1 & 30.1 & 8.3\\
        $\beta=0.5$ & 37.5 & 57.6 & 53.3 & 46.4 & 33.5 & 11.3\\
        \midrule
        $K_\beta=15$ & 37.6 & 57.3 & 53.1 & 46.0 & 33.9 & 12.5\\
        $K_\beta=10$ & 38.0 & 57.6 & 53.5 & 46.7 & 34.4 & 13.2\\
        $K_\beta=8$ & 37.6 & 57.5 & 53.3 & 45.9 & 33.9 & 12.4\\
        \bottomrule
        \end{tabular}
        \label{tab:dynamic_smoothL1}
    \end{minipage}
    \begin{minipage}{0.42\textwidth}
    \centering
        \makeatletter\def\@captype{table}\makeatother\caption{Inference speed comparisons using ResNet-50-FPN backbone on RTX 2080TI GPU.}
        \setlength{\tabcolsep}{1pt}
        \begin{tabular}{cc}
        \toprule
        Method & FPS\\
        \midrule
        Dynamic R-CNN & 13.9\\
        Cascade R-CNN & 11.2\\
        Dynamic Mask R-CNN & 11.3\\
        Cascade Mask R-CNN & 7.3\\
        \bottomrule
        \end{tabular}
        \label{tab:speed}
    \end{minipage}
\end{minipage}

\subsection{Studies on the effect of hyperparameters}

\textbf{Ablation study on $K_I$ in DLA.}
Experimental results on different $K_I$ are shown in Table~\ref{tab:dynamic_label_assignment}. Compared to the baseline, DLA can achieve consistent gains in AP regardless of the choice of $K_I$. These results prove the universality of $K_I$. Moreover, the performance on various metrics are changed under different $K_I$. Choosing $K_I$ as 64/75/100 means that nearly 12.5\%/15\%/20\% of the whole batch are selected as positives. Generally speaking, setting a smaller $K_I$ will increase the quality of selected samples, which will lead to better accuracy under higher metrics like $\mathrm{AP_{90}}$. On the contrary, adopting a larger $K_I$ will be more helpful for the metrics at lower IoU. Finally, we find that setting $K_I$ as 75 achieves the best trade-off and use it as the default value for further experiments. All these ablations prove the effectiveness and robustness of the DLA part.

\noindent
\textbf{Ablation study on $K_\beta$ in DSL.}
As shown in Table~\ref{tab:dynamic_smoothL1}, we first try different $\beta$ on Faster R-CNN and empirically find that a smaller $\beta$ leads to better performance. Then, experiments under different $K_\beta$ are provided to show the effects of $K_\beta$. Regardless of the certain value of $K_\beta$, DSL can achieve consistent improvements compared with various fine-tuned baselines. Specifically, with our best setting, DSL can bring 1.0 point higher AP than the baseline, and the improvement mainly lies in the high quality metrics like $\mathrm{AP_{90}}$ (+3.5 points). These experimental results prove that our DSL is effective in compensating for high quality samples and can lead to a better regressor due to the advanced dynamic design.

\noindent
\textbf{Ablation study on iteration count $C$.}
Due to the concern of robustness, we update $T_{now}$ and $\beta_{now}$ every $C$ iterations using the statistics in the last interval. To show the effects of different iteration count $C$, we try different values of $C$ on the proposed method. As shown in Table~\ref{tab:iteration_count}, setting $C$ as 20, 100 and 500 leads to very similar results, which proves the robustness to this hyperparameter.

\noindent
\textbf{Complexity and speed.}
As shown in Algorithm~\ref{algm:DynamicRCNN}, the main computational complexity of our method lies in the calculations of IoUs and regression labels, which are already done by the original method. Thus the additional overhead only lies in calculating the mean or median of a short vector, which basically \textbf{does not increase the training time}. Moreover, since our method only changes the training procedure, obviously the inference speed will not be slowed down.

Our advantage compared to other high quality detectors like Cascade R-CNN is the efficiency. Cascade R-CNN increases the training time and slows down the inference speed while our method does not. Specifically, as shown in Table~\ref{tab:speed}, Dynamic R-CNN achieves 13.9 FPS, which is $\sim$1.25 times faster than Cascade R-CNN (11.2 FPS) under ResNet-50-FPN backbone. Moreover, with larger heads, the cascade manner will further slow down the speed. Dynamic Mask R-CNN runs $\sim$1.5 times faster than Cascade Mask R-CNN. Note that the difference will be more apparent as the backbone gets smaller ($\sim$1.74 times faster, 13.6 FPS vs 7.8 FPS under ResNet-18 backbone with mask head), since the main overhead of Cascade R-CNN is the two additional headers.

\begin{table}[!t]
    \caption{The universality of Dynamic R-CNN. We apply the idea of dynamic training on Mask R-CNN under different backbones. ``bbox'' and ``segm'' stand for object detection and instance segmentation results on COCO \texttt{val} set, respectively.}
    \label{tab:universality}
    \begin{center}
\setlength{\tabcolsep}{3pt}
\begin{tabular}{cccccccc}
\toprule
Backbone & +Dynamic & $\mathrm{AP}^{bbox}$ & $\mathrm{AP}_{50}^{bbox}$ & $\mathrm{AP}_{75}^{bbox}$ & $\mathrm{AP}^{segm}$ & $\mathrm{AP}_{50}^{segm}$ & $\mathrm{AP}_{75}^{segm}$\\
\midrule
\multirow{ 2}{*}{ResNet-50-FPN} & & 37.5 & 58.0 & 40.7 & 33.8 & 54.6 & 36.0\\
 & \checkmark & 39.4 & 57.6 & 43.3 & 34.8 & 55.0 & 37.5\\
\midrule
\multirow{ 2}{*}{ResNet-101-FPN} & & 39.7 & 60.7 & 43.2 & 35.6 & 56.9 & 37.7\\
 & \checkmark & 41.8 & 60.4 & 45.8 & 36.7 & 57.5 & 39.4\\
\bottomrule
\end{tabular}
\end{center}

\end{table}

\subsection{Universality}

Since the viewpoint of dynamic training is a general concept, we believe that it can be adopted in different methods. To validate the universality, we further apply the dynamic design on Mask R-CNN with different backbones. As shown in Table~\ref{tab:universality}, adopting the dynamic design can not only bring $\sim$2.0 points higher box AP but also improve the instance segmentation results regardless of backbones. Note that we only adopt the DLA and DSL which are designed for object detection, so these results further demonstrate the universality and effectiveness of our dynamic design on improving training procedure for current detectors.

\begin{table}[!t]
    \caption{Comparisons of single-model results on COCO \texttt{test-dev} set.}
    \label{tab:overall}
    \begin{center}
\setlength{\tabcolsep}{3pt}
\begin{tabular}{lccccccc}
\toprule
Method & Backbone & $\mathrm{AP}$ & $\mathrm{AP}_{50}$ & $\mathrm{AP}_{75}$ & $\mathrm{AP_S}$ & $\mathrm{AP_M}$ & $\mathrm{AP_L}$\\
\midrule
RetinaNet~\cite{FocalLoss} & ResNet-101 & 39.1 & 59.1 & 42.3 & 21.8 & 42.7 & 50.2\\
CornerNet~\cite{CornerNet} & Hourglass-104 & 40.5 & 56.5 & 43.1 & 19.4 & 42.7 & 53.9\\
FCOS~\cite{FCOS} & ResNet-101 & 41.0 & 60.7 & 44.1 & 24.0 & 44.1 & 51.0\\
FreeAnchor~\cite{FreeAnchor} & ResNet-101 & 41.8 & 61.1 & 44.9 & 22.6 & 44.7 & 53.9\\
RepPoints~\cite{RepPoints} & ResNet-101-DCN & 45.0 & 66.1 & 49.0 & 26.6 & 48.6 & 57.5\\
CenterNet~\cite{CenterNet} & Hourglass-104 & 45.1 & 63.9 & 49.3 & 26.6 & 47.1 & 57.7\\
ATSS~\cite{ATSS} & ResNet-101-DCN & 46.3 & 64.7 & 50.4 & 27.7 & 49.8 & 58.4\\
\midrule
Faster R-CNN~\cite{FPN} & ResNet-101 & 36.2 & 59.1 & 39.0 & 18.2 & 39.0 & 48.2\\
Mask R-CNN~\cite{MaskRCNN} & ResNet-101 & 38.2 & 60.3 & 41.7 & 20.1 & 41.1 & 50.2\\
Regionlets~\cite{Regionlets} & ResNet-101 & 39.3 & 59.8 & - & 21.7 & 43.7 & 50.9\\
Libra R-CNN~\cite{LibraRCNN} & ResNet-101 & 41.1 & 62.1 & 44.7 & 23.4 & 43.7 & 52.5\\
Cascade R-CNN~\cite{CascadeRCNN} & ResNet-101 & 42.8 & 62.1 & 46.3 & 23.7 & 45.5 & 55.2\\
SNIP~\cite{SNIP} & ResNet-101-DCN & 44.4 & 66.2 & 49.9 & 27.3 & 47.4 & 56.9\\
DCNv2~\cite{DCNv2} & ResNet-101-DCN & 46.0 & 67.9 & 50.8 & 27.8 & 49.1 & 59.5\\
TridentNet~\cite{TridentNet} & ResNet-101-DCN & 48.4 & 69.7 & 53.5 & 31.8 & 51.3 & 60.3\\
\midrule
Dynamic R-CNN & ResNet-101 & 42.0 & 60.7 & 45.9 & 22.7 & 44.3 & 54.3\\
Dynamic R-CNN* & ResNet-101-DCN & 50.1 & 68.3 & 55.6 & 32.8 & 53.0 & 61.2\\
\bottomrule
\end{tabular}
\end{center}

\end{table}

\subsection{Comparison with State-of-the-Arts}

We compare Dynamic R-CNN with the state-of-the-art object detectors on COCO \texttt{test-dev} set in Table~\ref{tab:overall}. Considering that various backbones and training/testing settings are adopted by different detectors (including deformable convolutions~\cite{DCN,DCNv2}, image pyramid scheme~\cite{SNIP}, large-batch Batch Normalization~\cite{MegDet} and Soft-NMS~\cite{Soft-NMS}), we report the results of our method with two types.

Dynamic R-CNN applies our method on FPN-based Faster R-CNN with ResNet-101 as backbone, and it can achieve 42.0\% AP without bells and whistles. Dynamic R-CNN* adopts image pyramid scheme (multi-scale training and testing), deformable convolutions and Soft-NMS. It further improves the results to 50.1\% AP, outperforming all the previous detectors.

\section{Conclusion}

In this paper, we take a thorough analysis of the training process of detectors and find that the fixed scheme limits the overall performance. Based on the advanced dynamic viewpoint, we propose Dynamic R-CNN to better exploit the training procedure. With the help of the simple but effective components like Dynamic Label Assignment and Dynamic SmoothL1 Loss, Dynamic R-CNN brings significant improvements on the challenging COCO dataset with no extra cost. Extensive experiments with various detectors and backbones validate the universality and effectiveness of Dynamic R-CNN. We hope that this dynamic viewpoint can inspire further researches in the future.

\noindent
\textbf{Acknowledgements} This work is partially supported by Natural Science Foundation of China (NSFC): 61876171 and 61976203, and Beijing Natural Science Foundation under Grant L182054.

%
%
\bibliographystyle{splncs04}
\bibliography{egbib}

\clearpage

\section{Appendix}

\subsection{Effectiveness on One-stage Detectors}

Considering that dynamic training is a general viewpoint, we also try to apply the dynamic design on one-stage object detectors. Specifically, we choose a representative one-stage detector RetinaNet~\cite{FocalLoss} to validate the effectiveness of our method.

\begin{table}[!h]
    \caption{Effectiveness of the dynamic design on RetinaNet with ResNet-50-FPN as backbone on COCO \texttt{val} set.}
    \label{tab:sup_retinanet}
    \begin{center}
\setlength{\tabcolsep}{4pt}
\begin{tabular}{ccccccc}
\toprule
Method & $\mathrm{AP}$ & $\mathrm{AP}_{50}$ & $\mathrm{AP}_{75}$ & $\mathrm{AP_S}$ & $\mathrm{AP_M}$ & $\mathrm{AP_L}$\\
\midrule
RetinaNet & 35.6 & 55.4 & 38.4 & 20.3 & 39.5 & 46.5\\
Dynamic RetinaNet & 36.3 & 55.5 & 38.8 & 20.7 & 39.9 & 47.5\\
\bottomrule
\end{tabular}
\end{center}

\end{table}

As shown in Table~\ref{tab:sup_retinanet}, our dynamic design brings 0.7 points higher box AP than the RetinaNet baseline. It should be noted that RetinaNet has already changed the $\beta$ of SmoothL1 Loss to a small value (0.11), so in our experiment, adjusting $\beta$ to a slightly smaller value (0.05 with DSL) has little effect. Moreover, since the input of RetinaNet is the pre-defined anchors, the distribution of input is relatively fixed. So the impact of DLA can be regarded as using a more reasonable IoU threshold (e.g. 0.55) for training. We believe our method can be applied to other one-stage detectors if their inputs are more dynamic, like Guided Anchoring~\cite{GuidedAnchor}.

\subsection{Experimental Results on PASCAL VOC dataset}

\begin{table}[!h]
    \caption{Experimental results using ResNet-50-FPN backbone on PASCAL VOC2007 \texttt{test} set.}
    \label{tab:sup_voc}
    \begin{center}
\setlength{\tabcolsep}{4pt}
\begin{tabular}{ccccccc}
\toprule
Method & $\mathrm{AP}$ & $\mathrm{AP}_{50}$ & $\mathrm{AP}_{60}$ & $\mathrm{AP}_{70}$ & $\mathrm{AP}_{80}$ & $\mathrm{AP}_{90}$\\
\midrule
Faster R-CNN & 47.2 & 76.9 & 71.0 & 59.9 & 39.2 & 8.1\\
Dynamic R-CNN & 48.7 & 76.9 & 71.5 & 60.9 & 42.1 & 11.6\\
\bottomrule
\end{tabular}
\end{center}

\end{table}

To further demonstrate the effectiveness of our method, we conduct experiments on PASCAL VOC~\cite{PASCAL-VOC} dataset with the same hyperparameters as on MS COCO dataset. We use the union of VOC2007 and VOC2012 \texttt{trainval} as the training set and report the results on the VOC2007 \texttt{test} set. As shown in Table~\ref{tab:sup_voc}, Dynamic R-CNN improves the Faster R-CNN baseline by 1.5 points $\mathrm{AP}$ and 3.5 points $\mathrm{AP_{90}}$. Thus we reach similar conclusions as from the results on COCO dataset, which validates the effectiveness and universality of Dynamic R-CNN.

\end{document}